\documentclass[11pt,a4paper]{article}
\usepackage[hyperref]{acl2020}
\usepackage{pbox}
\usepackage{times}
\usepackage[utf8]{inputenc}
\usepackage[english]{babel}
\newtheorem{theorem}{Theorem}
\newtheorem{lemma}[theorem]{Lemma}
\newtheorem{definition}[theorem]{Definition}
\usepackage{amssymb}
\usepackage{listings}
\usepackage{latexsym}

\usepackage{xcolor}
\usepackage{amsmath}
\usepackage{microtype}

\aclfinalcopy % Uncomment this line for the final submission
\usepackage{mathtools}
\title{Semantic Scaffolds for Pseudocode-to-Code Generation}

\author{Ruiqi Zhong \quad Mitchell Stern \quad Dan Klein \\
  Computer Science Division \\
 University of California, Berkeley \\
  \texttt{\{ruiqi-zhong,mitchell,klein\}@berkeley.edu} \\}
\date{}

\begin{document}
\maketitle
\begin{abstract}
We propose a method for program generation based on \textit{semantic scaffolds}, lightweight structures representing the high-level semantic and syntactic composition of a program.
By first searching over plausible scaffolds then using these as constraints for a beam search over programs, we 
achieve better coverage of the search space when compared with existing techniques.
We apply our hierarchical search method to the SPoC dataset for pseudocode-to-code generation, in which we are given line-level natural language pseudocode annotations and aim to produce a program satisfying execution-based test cases.
% Syntactic well-formedness of a program barely implies that it can execute correctly.
% We incorporate variable declaration and usage constraints while translating human instructions into programs and searching for candidates that executes correctly. 
% To search efficiently under this constraint, we propose hierarchical beam search.
% It first generates \textit{scaffolds}, the minimum dependency information between code pieces from each line, and then generates candidates independently for each line conditioned on the scaffolds.
% Compared to regular beam search, our method is more computationally efficient, better explores variations near the beginning of the program and leads to higher performance.
%By using semantic scaffolds during inference, we attain a top-100 accuracy of 55.1\%, representing a 10.4\% absolute improvement over the previous state-of-the-art and reaching 81\% of the oracle performance for our model. 
By using semantic scaffolds during inference, we achieve a 10\% absolute improvement in top-100 accuracy over the previous state-of-the-art.
Additionally, we require only 11 candidates to reach the top-3000 performance of the previous best approach when tested against unseen problems, demonstrating a substantial improvement in efficiency.
\end{abstract}

\section{Introduction}

Systems that can map from natural language descriptions of tasks or programs to executable code have the potential for great societal impact, helping to bridge the gap between non-expert users and basic automation or full-fledged software development.
Accordingly, this area of research has garnered significant interest in recent years, with systems being devised for the translation of natural language specifications into database queries \cite{wang2018execution}, if-then programs \cite{chen2016latent}, game elements \cite{ling2016latent}, and more.

While much of the prior work in executable semantic parsing involves short descriptions being mapped into single-line programs, some tasks have recently been proposed that involve multiple natural language utterances on the input side and full programs on the output side, often reaching tens of lines in length and including non-trivial state manipulation.
Examples include
the Magic the Gathering and Hearthstone datasets \cite{ling2016latent} derived from trading cards and Java or Python classes implementing their behavior in a game engine,
% the CoNaLa dataset consisting of natural language intents and code snippets mined from Stack Overflow \cite{yin2018mining},
the CONCODE dataset \cite{iyer2018mapping} consisting of Java documentation strings and method bodies,
% the Spider dataset consisting of natural language questions and SQL queries \cite{yu2018spider},
and the NAPS and SPoC datasets \cite{zavershynskyi2018naps,kulal2019spoc} consisting of pseudocode annotations and source code for programming competition problems. 
%Of these, only the latter group includes input-output test suites, allowing for denotation-based evaluation.

\begin{figure}
    \centering
    \includegraphics[scale=0.5]{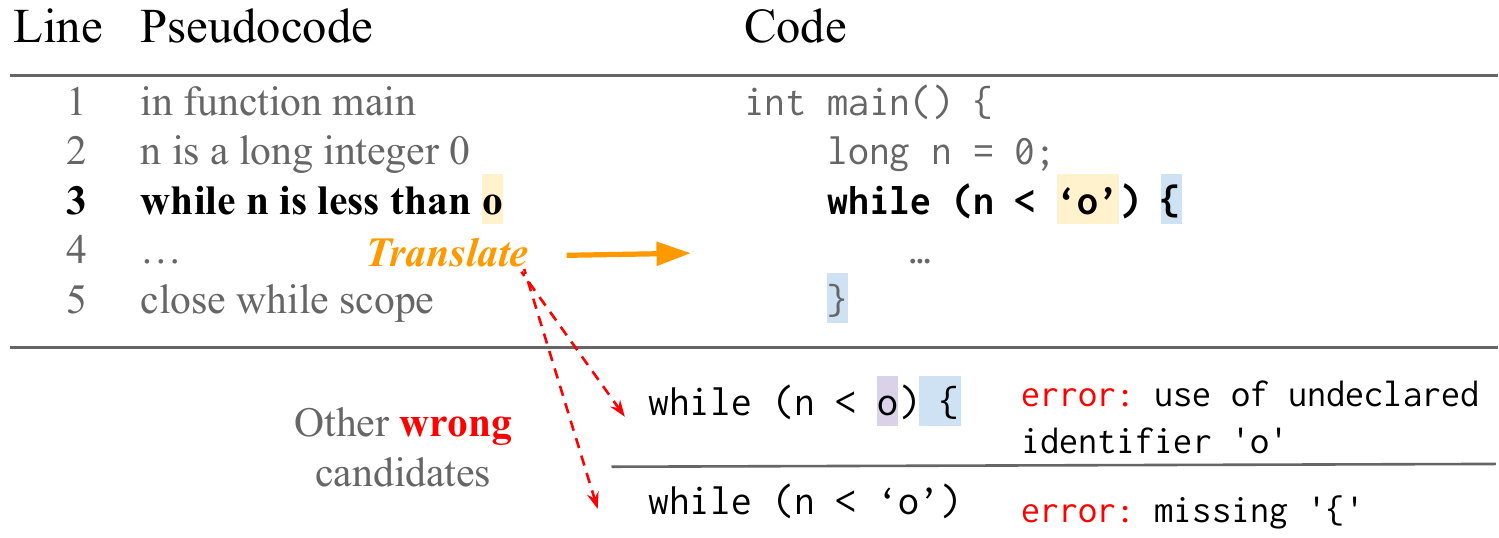}
    \caption{Pseudocode is translated to code for each line and combined to form a valid program. 
    Certain combinations are invalid due to syntactic and semantic constraints.}
    \label{fig:contraint-needed}
\end{figure}

Past approaches to these large-scale language-to-code tasks have typically employed sequence-based models \cite{ling2016latent} that do not account for structure on the output side, or tree-based models \cite{allamanis2015bimodal,rabinovich2017abstract,yin2017syntactic, hayati2018retrieval, iyer2019learning} that incorporate the syntax but not the semantics of the output domain. 
However, if we want to generate programs that can be executed successfully, the inclusion of both syntactic and semantic constraints is crucial.
As shown in Figure \ref{fig:contraint-needed}, while multiple program fragments may be syntactically correct and represent plausible translations of the corresponding pseudocode, not all of them will lead to executable programs.

To address this, we propose a search procedure based on \textit{semantic scaffolds}, lightweight summaries of higher-level program structure that include both syntactic information as well as semantic features such as variable declarations and scope constraints.
See Section~\ref{sec:combination-constraints} for a more formal definition.
While these do not encode the full spectrum of constraints used in some formal program synthesis tools \cite{solar2009sketching,gulwani2017program}, they strike a balance between utility, speed, and ease of use, offering substantial improvements in system performance without a significant increase in complexity.

% In this work we take into account semantic information such as variable scope, declaration and usage constraint while generating programs. 
% To be precise, each segment of code belongs to a variable scope; it cannot use a variable that has not been declared in a non-ending scope or repeatedly declare a variable that has been declared in the current scope.
% These constraints are fundamentally different from the syntactic Abstract Syntax Tree constraints, which have been intensively studied and widely employed by the current research community that translates human specifications into programs:
% provably, these constraints on variables cannot be specified by any Context Free Grammar by sub-exponential description size.

In this work we focus on the Search-based Pseudocode to Code (SPoC) dataset \cite{kulal2019spoc} due to its challenging multiline programs and availability of input-output test suites to  evaluate denotation accuracy.
The dataset contains line-level pseudocode annotations for 18,356 C++ programs provided by crowdsource workers from Amazon Mechanical Turk. 
As in the approach of \citet{kulal2019spoc}, we first obtain candidate code fragments for each line using an off-the-shelf neural machine translation system. 
We then aim to find the highest-scoring combination of fragments that results in a valid program. 
Although finding the optimal program under this setting is NP-hard when variable usage constraints are introduced (see Section \ref{hardness}), we can approximate it with a hierarchical beam search. 
Our algorithm first searches for semantic scaffolds for the program, then assembles fragments together conditioned on these scaffolds. 
This hierarchical approach speeds up search, produces higher quality variations, and leads to substantial improvements in our system's final accuracy.
% We apply this constraint to the Search based Pseudo code to Code (SPoC) dataset \cite{kulal2019spoc}.
% This is a dataset that contains ~20K programs with human annotated pseudo code (specification) on each line, and each program is accompanied with input-output test cases to examine whether the generation is semantically correct.
% To solve this dataset, pieces of code fragments are generated independently from each line's human specification, and we need to search for a valid solution that can pass the test cases by combining these code fragments together.
% Unfortunately, piecing these fragments together under variable constraint is in general NP-hard and hence we approximate the solution by beam search.
% This approximation is intolerably slow when beam width needs to be large and heavily biases variations at the end of the program.
% To fix this problem, we propose hierarchical beam search.
% The algorithm first searches for scaffolds - the minimum dependency between each line that includes information on syntactic components and variable usage/declarations. 
% Afterwards, candidates are selected independently conditioned on the scaffolds.

We achieve a new state-of-the-art by solving 55.1\% of the test cases within 100 attempts.
This represents a 10.4\% absolute improvement over the previous best \cite{kulal2019spoc}, and reaches 81\% of our model's oracle performance.
When tested against unseen problems (or crowd-workers), our top 11 (or top 52, respectively) candidates have the same performance as their top 3000 candidates, demonstrating marked gains in efficiency.

We complement our results with a discussion of specific cases in which our semantic scaffolds use global program context to resolve ambiguities in the pseudocode.
We also conduct a manual error analysis of 200 failures to better characterize the limitations of our method and suggest possible extensions for future work.

% In the final section we provide examples on what candidates are rejected and find that it helps disambiguate human instructions. 
% We also provide manual failure analysis on the code pieces generated by the model and hope that it can expedite future researches.

Our contributions are summarized as follows:
\begin{itemize}
\item We propose the use of semantic scaffolds to add semantic constraints to models for long-form language-to-code generation tasks.
\item We introduce a hierarchical beam search algorithm that incorporates these constraints, resulting in heightened efficiency, better coverage of the search space, and stronger performance when compared with the standard approach.
\item We achieve a new state-of-the-art accuracy of 55.1\% on the SPoC pseudocode-to-code dataset.
\end{itemize}

\section{Pseudocode-to-Code Task}

In this work, we focus on the SPoC dataset introduced by \citet{kulal2019spoc}.

\subsection{Data}
% All of our experiments are conducted on the SPoC dataset \cite{kulal2019spoc}.
This dataset consists of C++ solutions to problems from Codeforces, a competitive programming website, along with the input-output test cases used for each problem to evaluate correctness.
It contains 18,356 programs in total with 14.7 lines per program on average. 
Each line is annotated with a natural language pseudocode description given by a crowd worker from Amazon Mechanical Turk.
%An example is shown in Figure \ref{fig:example-data}.
On average, there are 7.86 tokens per line of code and 9.08 tokens per pseudocode annotation.
From the full dataset, 1,752 programs with annotations from unseen crowd workers and 1,820 programs for unseen problems are held out for evaluation.
More details can be found in \citet{kulal2019spoc}.
\iffalse
\begin{figure}
    \centering
    \includegraphics[scale=0.37]{plots/example-data.pdf}
    \caption{Example data with line pseudocode-to-code pairs. 
    Our goal is to translate pseudocode into executable program(s).
    }
    \label{fig:example-data}
\end{figure}
\fi 

\subsection{Task}
Suppose the target program has $L$ lines. For each line $l \in [L]$, we are given a natural language pseudocode annotation $x_{l}$ and an indentation level $i_{l}$. 
Our goal is to find a candidate program $y$ based on $(x_1,i_1), \dots, (x_L, i_L)$ that can solve the given problem (i.e. pass all the test cases) using as few submission attempts as possible. 
The search efficiency of an algorithm is calculated as the fraction of problems it can solve using a budget of $B$ attempts per problem, where an attempt includes both compiling a candidate program and running the test cases.

As in \citet{kulal2019spoc}, for each pseudocode line $x_{l}$, we use an off-the-shelf neural machine translation system to obtain a set of $C$ candidate code pieces $Y_{l} = \{y_{lc} \mid c \in [C]\}$, where candidate code piece $y_{lc}$ has probability $p_{lc}$.
A full candidate program $y$ is a concatenation of candidate code pieces, one per line, and has score $p(y)$:
\begin{equation}
    y = \mathrm{concat}_{l=1}^{L}y_{lc_{l}}, \qquad p(y) = \prod_{l=1}^{L}p_{lc_{l}} .
\end{equation}
We aim to find valid high-scoring programs in our search procedure.

\section{Combination Constraints}
\label{sec:combination-constraints}

\citet{kulal2019spoc} propose best-first search as a baseline, which enumerates all complete candidate programs in descending order by score.
Using a priority queue, this algorithm can efficiently find the exact top $B$ highest scoring candidates in time $O(L \log (BL))$ per candidate.

However, this approach ignores any dependence between different lines. 
For example, any of the code piece candidates in Figure~\ref{fig:contraint-needed} could potentially be used in a valid program, but if we naively combine certain subsets of candidates together, the resulting program will be invalid due to the use of undeclared variables or mismatching braces.
To solve this problem, we propose to enforce certain \textbf{syntactic} and \textbf{semantic} constraints when combining candidate code pieces.

\subsection{Syntactic Constraints} \label{sec:syntactic-constraints}

\begin{figure*}
    \centering
    \includegraphics[scale=0.33]{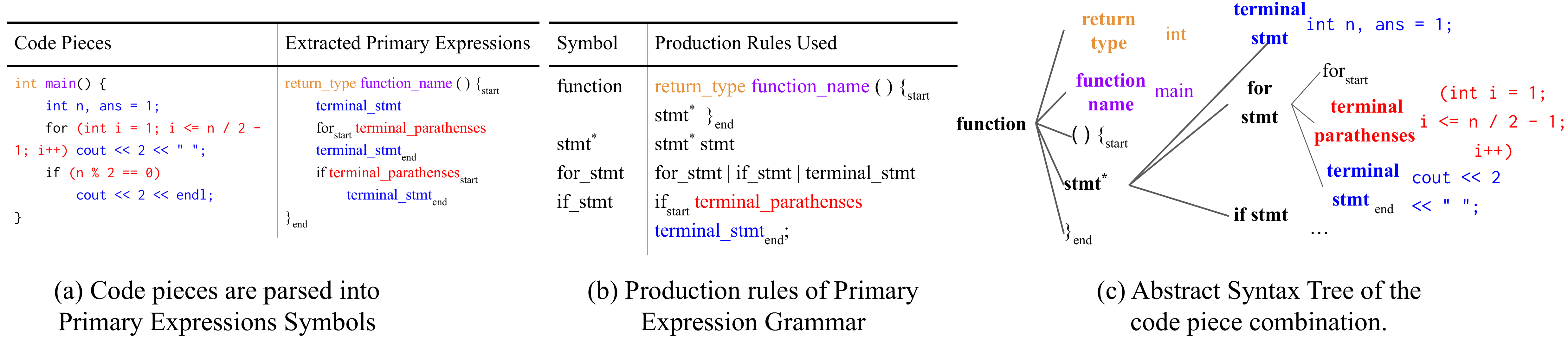}
    \caption{Example primary expression grammar. Subscripts ``start/end" refers to starting/ending variable scopes.
    }
    \label{fig:primary_expression}
\end{figure*}

The candidate program should adhere to the grammatical specification of the target language. % \cite{aho1986compilers}.
However, since incorporating the complete set of C++ grammatical constraints would require significant engineering effort, we instead restrict our attention to the set of ``primary expressions" consisting of high-level control structures such as \texttt{if}, \texttt{else}, \texttt{for} loops, function declarations, etc.
%and assume that the lower level statements are syntactically valid.
As shown in Figure \ref{fig:primary_expression}, we parse the candidate code pieces for each line into a list of primary expression symbols.
In order for code pieces from consecutive lines to be used together, there must exist a grammatical derivation that combines their respective symbols.
The complete list of primary expression can be found in the appendix; see Tables \ref{tab:primary_expression_table} and \ref{tab:terminals}.

Additionally, some production rules are associated with the start or end of a variable scope block.
We require that the number of open scope blocks equals the indentation level $i_{l}$ for each line $l$.

\subsection{Symbol Table Constraints} \label{sec:symTable-constraints}
\begin{figure}
    \centering
    \includegraphics[scale=0.5]{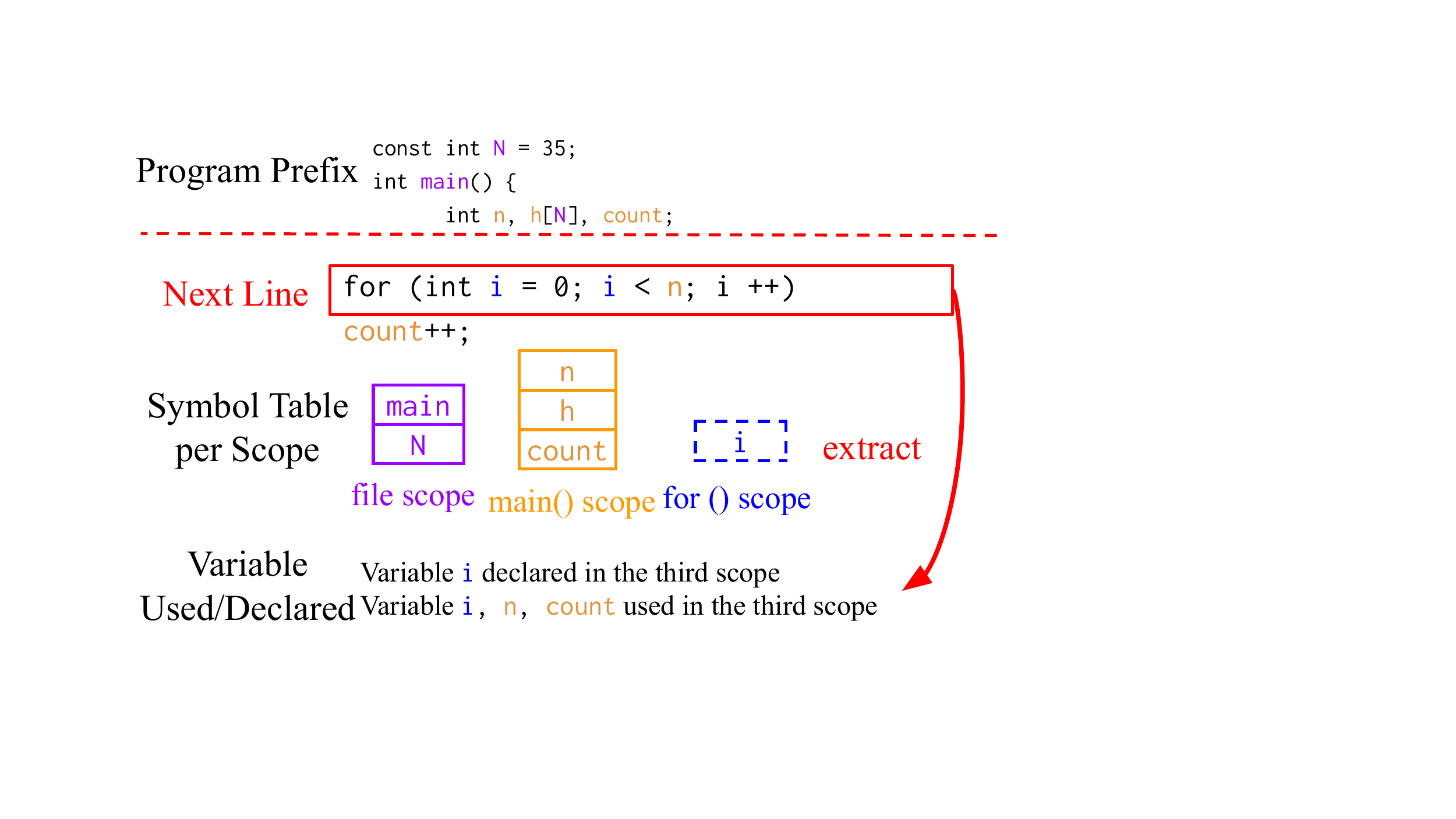}
    \caption{Extracting variables used or declared at each scope for a given code piece to verify the symbol table constraints.}
    \label{fig:variabe_used}
\end{figure}

Each scope block is associated with a symbol table \cite{aho1986compilers} keeping track of the variables that have been declared within that scope or any containing scopes.
% We assume that each code piece candidate belongs to one variable scope and
We extract the variable names used or declared by each code piece (Figure \ref{fig:variabe_used}) and ensure that (1) undeclared variables are not used, and (2) variables are not redeclared within the same scope.
% When selecting code pieces, we ensure that a code piece (1) cannot use a variable that is not in any of the symbol tables associated with a non-ending variable scope and 2) cannot declare a variable that has already been in the symbol table of the current variable scope.
After checking these constraints, any variables declared by a given code piece will be added to the symbol table associated with the current scope. 
% Put in simple words, code pieces cannot use a variable that is undeclared or repeatedly declare a variable in the same scope.

These symbol table constraints are based on the \textbf{semantic} information of code pieces and are fundamentally different from previous AST-based syntactic constraints for code generation \cite{rabinovich-etal-2017-abstract,yin2017syntactic}.
Formally, 
%even if we know the variable names when we create the production rules, 
any context free grammar that specifies the same constraints requires at least exponential description complexity.
We provide a proof adapted from \citet{ellul2005regular} in Appendix~\ref{sec:CFGDescription}.

\subsection{Syntactic and Semantic Scaffolds}
We note two properties of the aforementioned constraints.
First, we can efficiently compute whether a program prefix can possibly lead to a full program that satisfies the constraints by using an incremental parser \cite{ghezzi1979incremental} and checking the symbol tables.
Secondly, not all information from a code piece is necessary to verify the constraints.
Accordingly, when multiple code piece candidates have the same primary expression symbols and variable declarations and usage, swapping between them would not affect the satisfiability of the constraints.
For example, changing from \texttt{a += 1} to \texttt{a -= 1} will not change a compilable program into a non-compilable one, or vice versa.
These two properties will help motivate the hierarchical beam search algorithm introduced in the next section.

More formally,
% define a non-injective function $\phi(y_{lc})$ to be the \emph{configuration} of line $y_{lc}$ that contains all the information needed about it to verify the constraints.
we take the \textit{configuration} $\phi(y_{lc})$ of a line $y_{lc}$ to be the minimal set of features required to verify the above constraints.
The \textit{prefix scaffold} $S_{y,l} = [\phi(y_{1c_{1}}), \phi(y_{2c_{2}}), \dots, \phi(y_{lc_{l}})]$ of a program $y$ then contains all the information needed to verify the constraints for the first $l$ lines.
%, and is defined as:
%\begin{equation}
%    S_{y, l} = [\phi(y_{1c_{1}}), \phi(y_{2c_{2}}) \dots \phi(y_{lc_{l}})]
%\end{equation}
We can efficiently compute whether $S_{y, l}$\footnote{To keep notation uncluttered, we sometimes use $\phi$ to denote a configuration, we ignore the subscript $y$ of $S$ when we refer to a general scaffold that is not necessarily associated with a specific program, and we ignore the subscript $l = L$ of $S$ when we refer to the scaffold of a full program.} is a valid prefix scaffold when $l < L$ and whether $S_{y, L}$ is a valid scaffold for a full program when $l = L$.

\section{Constrained Search}
\subsection{Beam Search}
Our goal is to find the top $B$ highest-scoring candidate programs that satisfy the aforementioned constraints. 
Unfortunately, finding whether even one solution exists is NP-hard (proof given in Section \ref{hardness}).
One way we can approximate the solution is to use a standard beam search.
The beam maintains a list of hypothesis program prefixes along with their respective scores.
We extend the beam by adding the candidate code pieces from the next line to each candidate program prefix if they form valid combinations under the constraints, then prune the hypotheses with scores outside of the top $W$.
The algorithm ends after $L$ steps, returning all the valid hypotheses in the final beam.

\subsection{Scaffold Search}
Although beam search can approximate the top $B$ solutions, the time complexity of beam search grows quadratically with the beam width $W$. 
Finding the top $B$ candidates requires that $W \geq B$, and hence each candidate takes $\Omega(BL)$ (amortized) time to generate, which can become intractable if $B$ is on the order of thousands.
Even worse, beam search is often biased towards variations at the end of the program due to its greedy decisions, and can waste its budget on candidates that are unlikely to be the correct solution.

This is in direct contrast to the computationally lighter baseline which generates the exact (unbiased) top candidates independently for each line without constraint. 
Can we combine the advantages of both algorithms?
A key observation is that the assumption of \textit{independent} scoring across different lines allows fast and unbiased full program candidate generation, while an expensive beam search is inevitably needed to deal with the inherent \textit{dependence} between lines.

Therefore, we propose a hierarchical beam search method that first uses beam search with a smaller beam width $W$ to find likely scaffolds, including only the minimum dependency information between lines to satisfy the constraints, then scores candidates independently for each line conditioned on the scaffold. 
We assign probability $p(\phi_{l\gamma})$ to configuration $\phi_{l\gamma}$  by marginalizing all code piece candidates at line $l$ with configuration $\phi_{l\gamma}$, and assign probability $p(S)$ to scaffold $S$ by multiplying the configuration probabilities from each line:
\begin{equation}
    p(\phi_{l\gamma}) = \sum_{\phi(y_{lc}) = \phi_{l\gamma}}p_{lc}, \qquad p(S) = \prod_{i=1}^{L}p(S[i]) .
\end{equation}
Using this scoring function, we run a scaffold beam search with size $W$, then select the top $K$ highest scoring scaffolds $S_{1}, S_{2} \dots S_{K}$.

Next, to generate program candidates from a given scaffold $S$, we filter out all code pieces in $Y_{l}$ that do not have the configuration specified by $S$; in other words, the new set of code candidate pieces for each line $l$ is
\begin{equation}
    Y_{l}^{S} = \{y_{lc} \in Y_{l} \mid \phi(y_{lc}) = S[l]\} .
\end{equation}
As a result, conditioned on a fixed scaffold $S$, code pieces from each line can be chosen independently and the resulting full program will be \textit{guaranteed} to satisfy the aforementioned constraints.
%Figure \ref{fig:scaffold} illustrates our algorithm.

Given $K$ candidate scaffolds, we enumerate the top full program candidate from each scaffold and choose the highest scoring one.
This takes time $O(K + L\log (BL))$ per candidate.
In practice, we pick relatively small $K$ and the running time has only logarithmic dependence on $B$.

%\footnote{rigorously speaking, we got rid of polynomial dependence on $B$, since there is a $\log B$ factor incurred by the priority queue.}.

\begin{figure*}
    \centering
    \includegraphics[scale=0.61]{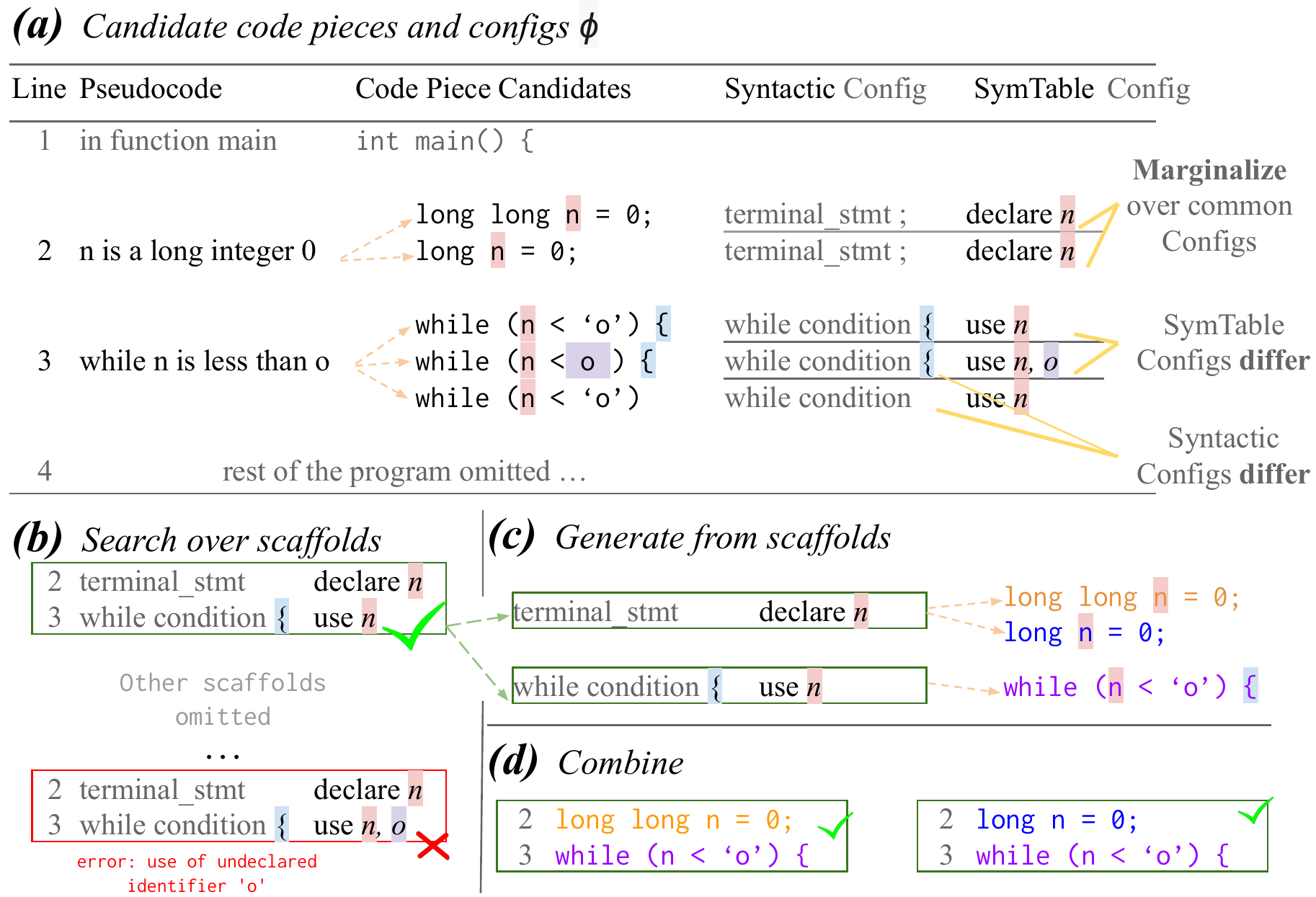}
    \caption{
    (a) Candidate code pieces and their syntactic/Symtable configuration for each line; 
    (b) use beam search to find highest scoring valid scaffolds;
    (c) given a scaffold, select code pieces that has the same configurations for each line.
    (d) combine code pieces to form full program.
    }
    \label{fig:scaffold}
\end{figure*}

\subsection{Tradeoffs in Early Detection} \label{brute-force}
An alternative view on beam search is that it front loads the computation to reject invalid programs that do not satisfy the constraints earlier in the search process.
A brute force alternative is to generate the next highest scoring candidates from the unconstrained baseline and reject invalid ones.
This method is guaranteed to produce top-scoring solutions, but it might need arbitrarily many candidates to find a valid one.
We need to compare the computational efficiency between these two methods.

The most computationally expensive operation in constraint verification is to verify whether the next line is valid given the program prefix.
Therefore, we count how many times this verifier function is called as a proxy to measure computational efficiency.
We allow the brute force method to use as large a verifier function call quota as our ``active" beam search method: it can validate/reject a program candidate until the quota is used up.

Section \ref{brute-force-comparison} compares our scaffold search method against this brute force approach.
The latter needs thousands of times more computation to attain the same level of performance as the former.

\section{Implementation\footnote{Our implementation is available at \url{https://github.com/ruiqi-zhong/SemanticScaffold}.}}\label{sec:implementation}

\paragraph{Empty Pseudocode} Around 26\% of the lines in the data set do not have pseudocode annotations. 
They usually correspond to lines of code that do not have semantically meaningful information, such as ``\texttt{int main() \{}'',  ``\texttt{\{}'', ``\texttt{\}}'', etc. 
\citet{kulal2019spoc} replaced these empty pseudocode lines with the ground truth code, effectively giving this information away to the search algorithm.
We did not use the gold code pieces for these lines, which makes our task more challenging.

\paragraph{Model Training}  
We use OpenNMT \cite{2017opennmt} with its default settings to translate pseudocode into code piece candidates. Our model is a two-layer LSTM seq2seq model with hidden size 512, an attention mechanism \cite{bahdanau2014neural} and copy pointers \cite{vinyals2015pointer}.

We estimate the fraction problems solvable given infinite search budget and 100 candidates per line as in \citet{kulal2019spoc} to obtain an oracle bound on performance.
Due to slight difference in hyperparameters and tokenization method, our model has higher ceiling: on the unseen worker (problems) test set, the oracle performance\footnote{The oracle performance here is not a universal property of the data, but depends on the model used to generate the code pieces.} is 74.4\% (60.5\%), compared to 71.4\% (55.2\%) in previous work.
Across all test examples, the oracle performance is 68\%.

\paragraph{Parsing Code Pieces}
Since no off-the-shelf C++ parser extracts the information we need from code pieces, 
%and hacking them for our own purposes requires significant amount of engineering, 
we implement our own primary expression parser to extract high level control information. 
We rely on the following heuristic assumptions to parse the code pieces generated by the model: (1) a code piece belongs to only one variable scope; (2) the generation of every primary expression terminal symbol lies in one line.
Our parser fails on less than $0.01\%$ of the code pieces in the dataset. 
While selecting the candidates for each line, we immediately reject the ungrammatical pieces we cannot parse.
Without deliberate implementation optimization, this parsing operation takes on average 2.6 seconds to process all the top 100 code pieces for a problem -- approximately the same wallclock time as 1 compilation attempt.

\paragraph{Search Algorithm Hyperparameters}
As in \citet{kulal2019spoc}, we consider the top $C=100$ code pieces for each line. 
Unless otherwise mentioned, our default beam width $W$ is 50 for scaffold search and we keep the top $K = 20$ scaffolds for the subsequent generation.

\section{Search Performance}
\subsection{Metrics}
We evaluate a search algorithm $\mathcal{A}$ by computing the fraction of problem it can solve on the test set given evaluation budget $B$ per problem, which we denote as $f_{\mathcal{A}}(B)$.
%\footnote{As a sanity check, $f_{\mathcal{A}}$ is an increasing function.}
%, since the more candidate program an algorithm generates and attempts, the more fraction of programs it can solve.
We plot $f_{\mathcal{A}}$ against $B$ and evaluate it at $B=1, 10, 100, 1000$ for each algorithm $\mathcal{A}$ to compare performance.

We note that the difference of $f$ values between two algorithms becomes smaller and less informative as $B$ increases. 
With infinite code piece candidates and budget, a brute force search can enumerate all possible programs, find the right solution and $f$ converges to 1. 
Direct comparison on $f$ values hence becomes meaningless as $B$ increases.
To address this deficiency, we define a \textit{lead} metric $l_{\mathcal{A}_{1}, \mathcal{A}_{2}}(B)$ equal to the extra budget $X$ needed by algorithm $\mathcal{A}_{2}$ to reach the same level of performance as $\mathcal{A}_{1}$ given budget $B$.
Formally,
\begin{equation}
    l_{\mathcal{A}_{1}, \mathcal{A}_{2}}(B) = \inf \{X \mid f_{\mathcal{A}_{2}}(B + X) \geq f_{\mathcal{A}_{1}}(B)\} .
\end{equation}
A visualization can be seen in Figure \ref{fig:main}(c).

We report our algorithms' performance on the heldout test set with annotations from unseen crowd workers and with unseen problems separately.
%We present two sets of results here: 1) comparison of different type of constraints and their effects on search performance and 2) comparison of hierarchical beam search vs. regular beam search with various sizes.

\subsection{Comparison of Constraints}

We compare four settings:
\begin{itemize}
    \item \textbf{No Constraints}: the best-first search method that scores lines independently.
    \item \textbf{Syntactic Constraints}: the constraints on the primary expression and indentation level as described in section \ref{sec:syntactic-constraints}. 
    %Correspondingly, the configurations $\phi$ we extract do not contain variable %usage/declaration information.
    \item \textbf{Symbol Table Constraints}: both the syntactic constraints and the symbol table constraints described in section \ref{sec:symTable-constraints}. 
    We abbreviate this as \textbf{SymTable}. 
    \item \textbf{Backoff}: sometimes hierachical beam search with the SymTable constraints fails to return any valid scaffold. 
    We back off to just the Syntactic constraints if this happens.
\end{itemize}
Additionally, we compare with the \textbf{Previous} state-of-the-art reported by \citet{kulal2019spoc}.
%We do not run $B = 3000$ due to constraint on computational resources. 
\begin{figure}
    \centering
    \includegraphics[scale=0.16]{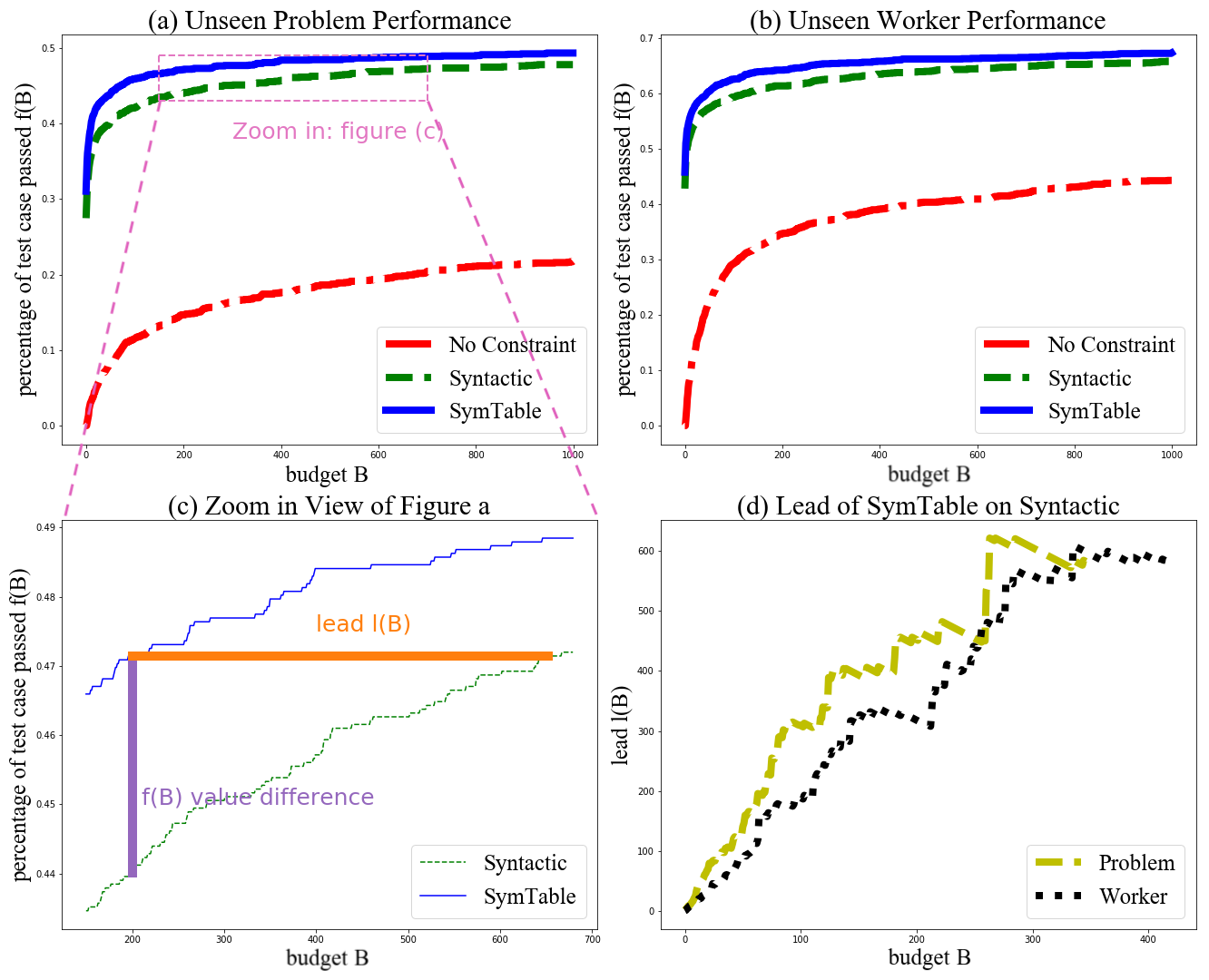}
    \caption{(a), (b) Comparison of $f$ performance under different constraints. (c) a zoom in visualization on the definition of \textit{lead} metrics (d) \textit{lead} of SymTable constraint on Syntactic constraint on different test sets.}
    \label{fig:main}
\end{figure}

The results can be seen in Figure \ref{fig:main} and Table \ref{tab:main_result}, where we use the constraint type as a shorthand for the search algorithm under this constraint.
Without constraints, the baseline algorithm performs especially poorly because it needs syntactic context to select relevant code pieces for $26\%$ of the lines with empty pseudocode. 

SymTable outperforms Syntactic.
As shown in Figure \ref{fig:main}(d), the lead of SymTable on Syntactic grows linearly: the more these two algorithms search, the more budget is needed by Syntactic to reach the same level as SymTable.
Syntactic needs nearly 600 more budget to have comparable performance with SymTable that uses 400 budget.
%We observe these trends consistently and robustly across different budget limit and test sets.

We notice that all of our constrained search methods outperform the previous state-of-the-art.
Averaged across all test examples, Backoff can solve 55.1\% of the problems within 100 budget, which is $\approx10\%$ higher than the previous work.
On unseen workers (problems), the top 11 (top 52) candidates of Backoff solve the same fraction of problems as the top 3000 candidates of the best performing algorithm in \citet{kulal2019spoc}.
%Nevertheless, we emphasize that our search algorithms are not \textit{directly comparable}: 1) our seq2seq model oracle accuracy is slightly higher than previously reported  and 2) \citet{kulal2019spoc} uses gold code piece for empty pseudocode.

\begin{table}[]
    \centering
    \textbf{
    Test Against Unseen Workers}\\
    Hierarchical Search (\textbf{H}), Beam Width \textbf{W = 50}
    \begin{tabular}{ccccc}
    \hline
        Constraint & $B$=1 & $B$=10 & $B$=10$^{2}$ & $B$=10$^{3}$  \\
        \hline
        None & 0.0\% & 8.1 \% & 29.2 \% & 44.3\%\\
        Previous & 30.7\% & 44.4\% & 53.7\% & 58.6\% \\
        \colorbox{blue!30}{Syntactic} & 42.8 \% & 51.9\% & 59.3\% & 65.9\%\\
        \colorbox{red!30}{SymTable} & 45.8\% & 55.1\% & 62.6\% & 67.3\%\\
        Backoff & \textbf{46.0\%} & \textbf{55.3}\% & \textbf{62.8\%} & \textbf{67.6\%}\\
        \hline
    \end{tabular}
    \textbf{Test Against Unseen Problems}
    \begin{tabular}{ccccc}
    \hline
        Constraint & $B$=1 & $B$=10 & $B$=10$^{2}$ & $B$=10$^{3}$  \\
        \hline
        None & 0.0\% & 3.0\% & 11.5\% & 21.8\%\\
        Previous & 17.8\% & 28.4\% & 34.2\% & 38.3\% \\
        Syntactic & 27.5 \% & 35.4\% & 42.1\% & 47.8\%\\
        SymTable & 31.0\% & 39.2 & 46.0\% & 49.3\%\\
        Backoff & \textbf{31.2\%} & \textbf{39.4\%} & \textbf{46.1\%} & \textbf{49.6\%} \\
        \hline
    \end{tabular}
    \caption{Comparison of the fraction of program passed when $B=10^{0, 1, 2, 3}$ under different constraints; constraint satisfied by hierarchical beam search with the default hyper-parameters mentioned in Section \ref{sec:implementation}.
    ``Previous" refers to the previous state of the art model.
    }
    \label{tab:main_result}
\end{table}

\subsection{Regular vs. Hierarchical Beam Search}
We use regular beam search with beam width $W = 200$ to generate $B = 100$ valid candidate full programs. 
We did not experiment with $B = 1000$ because beam search with $W \geq B \geq 1000$ is computationally intractable.
For hierarchical beam search we experiment with $W = 10, 25, 50$ for scaffold search and keep the top $K = min(W, 20)$ scaffolds for subsequent searches. 

Table \ref{tab:worker_beamsize} compares the performance of hierarchical beam search against regular beam search with different beam sizes under Syntactic and SymTable constraints. 
We find that if hierarchical beam search is used, even dropping the beam width from 50 to 10 leads to negligible change in performance.
In contrast, even with a large beam width $W=200$, regular beam search method cannot efficiently search for the solution and leads to a noticeable drop in performance.
%, even though it is more computationally expensive than hierarchical search with $W = 10$. 

We observe a similar trend for SymTable: regular beam search with beam width $W=200$ under-performs hierarchical search with beam width $W=25$.
However, if we further decrease the hierarchical beam search width from 25 to 10 in this setting, we observe a significant drop in performance, possibly because there are more variable usage variations than syntactic variations.

\subsection{Scaffold Search vs. Brute Force Method} \label{brute-force-comparison}
We now compare scaffold search to the brute force algorithm as described in section \ref{brute-force}.
We make $B = \text{50,000}$ attempts for the brute force method so that its performance can match at least the top 10 candidates of our constrained approach and make the lead metrics  meaningful. 
To save computation and avoid compiling all 50,000 programs, we early reject every candidate that does not fulfill our constraints.

\begin{figure}
    \centering
    \includegraphics[scale=0.17]{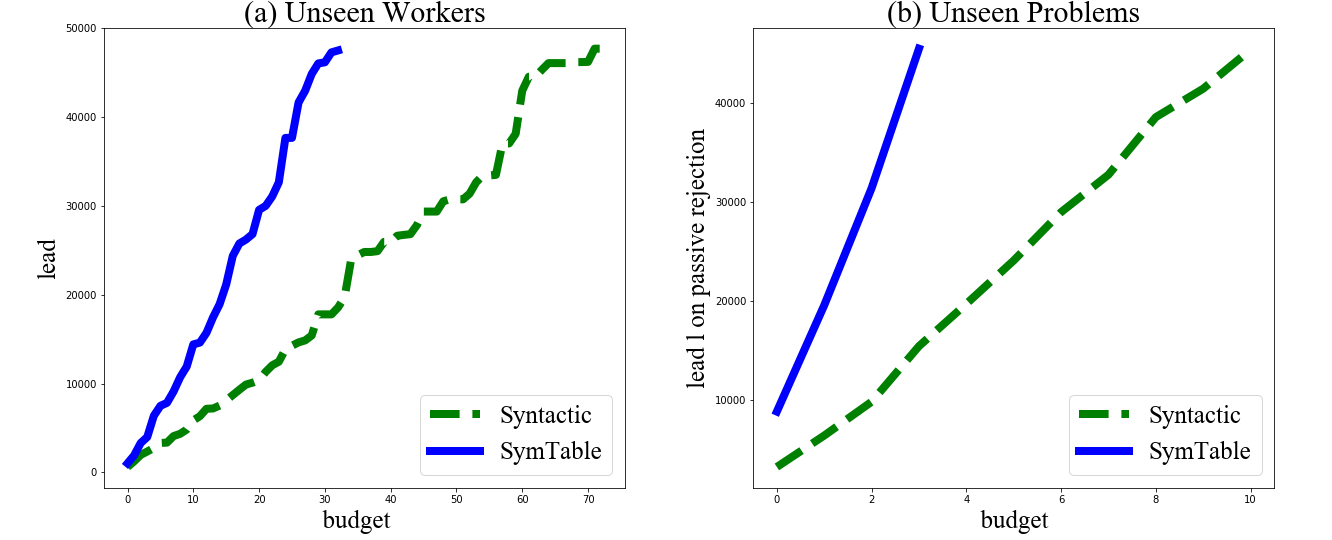}
    \caption{Lead of SymTable and Syntactic constraints on non-constrained approach with equal quota on test set with unseen (a) workers and (b) problems.}
    \label{fig:lead_on_passive}
\end{figure}

The lead of our approaches against the brute force algorithm is shown in Figure \ref{fig:lead_on_passive}.
After being adjusted for the constraint checking quota used, the lead of our approach is tens of thousands ahead of the unconstrained approach.
Scaffold search saves lot of computation by inducing a little overhead earlier in the search process. 

\begin{table}[]
    \centering
    \textbf{Test Against Unseen Workers, Syntactic}
    \begin{tabular}{cccc}
    \hline
        Method, Width & $B$=1 & $B$=10 & $B$=10$^{2}$  \\
        \hline
        H, $W$=10 & 42.8\% & 51.7\% & 59.1\% \\
        H, $W$=25 & 42.8\% & 51.8\% & 59.3\%\\
        \colorbox{blue!30}{\textbf{H,} $\mathbf{W}=50$} & \textbf{42.8\%} & \textbf{51.9\%} & \textbf{59.3\%}\\
        R, $W$=200 & 42.4\% & 51.3\% & 58.2\%\\
        
        \hline
    \end{tabular}
    
    \textbf{Test Against Unseen Workers, SymTable}
    \begin{tabular}{cccc}
    \hline
        Method, Width & $B$=1 & $B$=10 & $B$=10$^{2}$  \\
        \hline
        H, $W$=10 & 45.4\% & 54.3\% & 61.0\% \\
        H, $W$=25 & 45.6\% & 54.7\% & 61.9\%\\
        \colorbox{red!30}{\textbf{H,} $\mathbf{W}=50$}& \textbf{45.8\%} & \textbf{55.1\%} & \textbf{62.6\%}\\
        R, $W$=200 & 45.6\% & 54.9\% & 61.9\%\\
        
        \hline
    \end{tabular}
    
    \caption{Comparison of different beam size with Syntactic and SymTable constraint when tested against unseen workers. H/R refers to hierarchical/regular beam search and $W$ is the beam width.
    The same results on unseen problems can be seen in appendix .
    }
    \label{tab:worker_beamsize}
\end{table}

\section{Analysis}
\begin{figure*}
    \centering
    \includegraphics[scale=0.64]{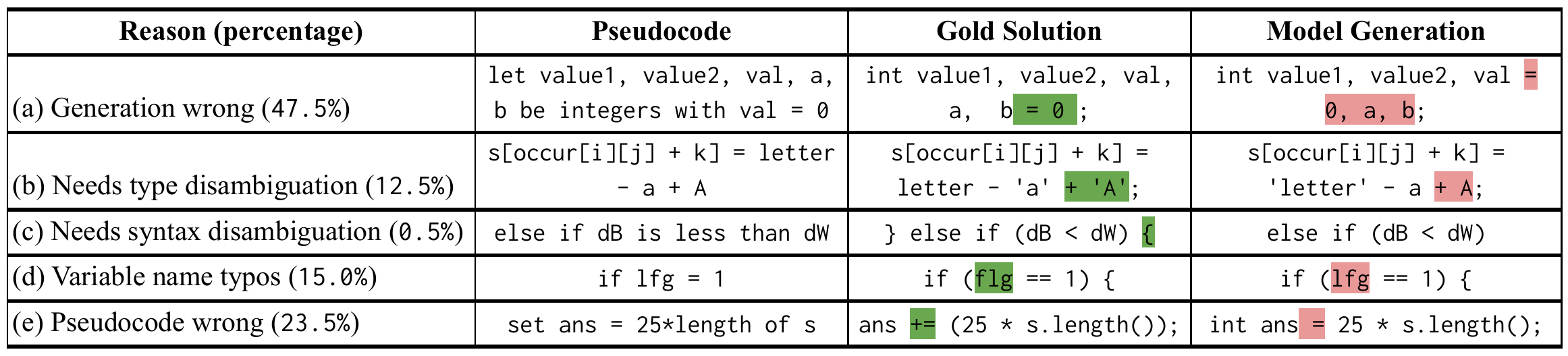}
    \caption{Categorized error analysis for lines that no generated code piece is functionally equivalent to the gold. The percentage in the parentheses refers to the fraction of this category out of the 200 samples.}
    \label{fig:categorized_error_analysis}
\end{figure*}

\begin{figure}
    \centering
    \includegraphics[scale=0.5]{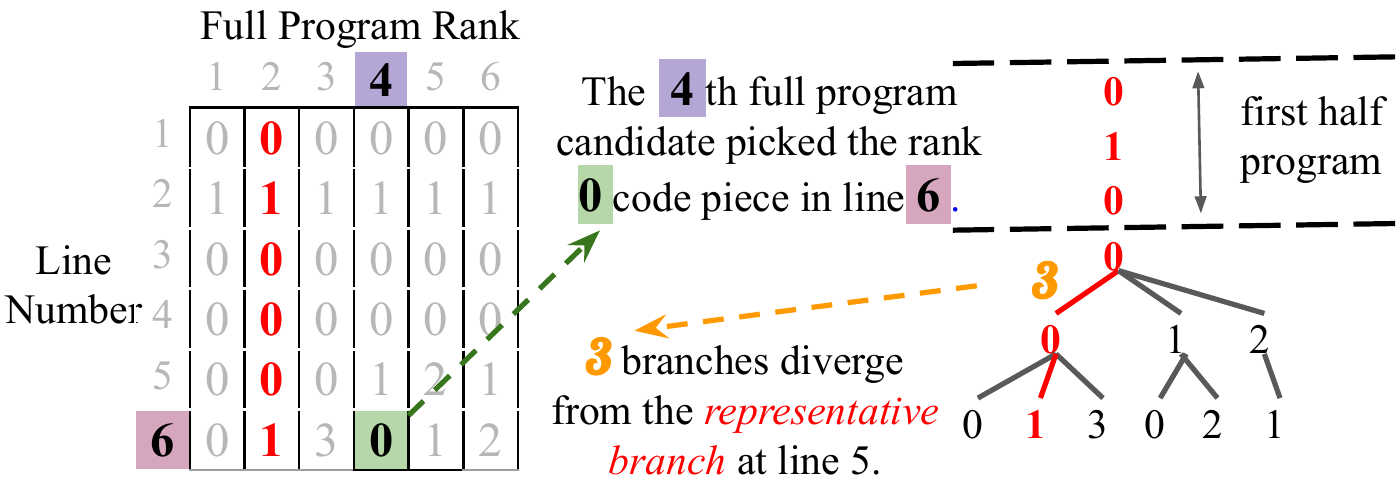}
    \caption{(a) A matrix that represents each candidate's choices of code pieces for each line.
    (b) A prefix tree constructed by treating each column as a string; the \textit{representative branch} is the second column and marked with red color. 
    } 
    \label{fig:candidate_variation}
\end{figure}

\subsection{Program Candidate Variations} \label{beamvariation}
Beam search has the problem of producing fewer variations at the beginning of the search.
Such a weakness might be tolerable if we only care about the top 1 candidate, but becomes disastrous in a search setting where we want the top $B$ candidates, whose variation is typically spread across the entire program.

We describe the following procedure to formally define this intuition. 
We first aggregate code piece choices for each line for all the top $B$ programs.
As shown in Figure \ref{fig:candidate_variation}(a), we construct a matrix such that each column corresponds to a full program candidate; 
the number $r$ in the $i^{th}$ row and $j^{th}$ column means that on line $i$, the $j^{th}$ full program candidate chooses the $r^{th}$ code piece candidate (i.e. $y_{ic_{i}} = y_{ir}$).
Then we can build a prefix tree (Figure \ref{fig:candidate_variation}(b)) by treating each column as a string, where each traversal from the root to a leaf is a complete candidate program $y$.
We define the \textit{representative branch}/program as a traversal from the root to a leaf that always chooses the child that contains the most leaves (with ties being broken randomly).
For each of the remaining $B - 1$ programs/traversals, we find the smallest line number where it starts to diverge from the representative branch.
Among these $B - 1$ programs, we count the fraction of divergences that take place in the first/second half of the lines.
For example, in Figure \ref{fig:candidate_variation}(b), $0\%$ of the divergences occur in the first half.

We compare hierarchical vs.\ regular beam search under syntactic constraints with different beam widths $W$: hierarchical $W = 10, 50$ and regular $W = 50, 200$.
We group the programs by length $L$, consider the top $B = 25$ attempted programs for each problem and report the fraction of divergences that occur in the first half of the program length for each group.

The results can be seen in Table \ref{tab:candidatevariation}.
For regular beam search, a moderate beam width $W=50$ consistently brings fewer variations in the first half of the program, and it needs a larger $W=200$ to fix this problem.
In contrast, a small $W$ for hierarchical beam search produces the same amount of variations in the first half of the program.
The same statistics under SymTable constraints can be seen in the appendix (Table \ref{tab:candidatevariation_SymTable}) and the conclusion holds similarly.

\begin{table}[]
    \centering
    \begin{tabular}{c|cccc}
    \hline
    Length $L$ & H 10 & H 50 & R 50 & R 200\\
      \hline
    (0, 10] & 45.4\% & 45.5\% & \underline{43.6\%} & 45.5\%  \\
    (10, 20] & 63.2\% & 63.4\% & \underline{58.2\% }& 63.4\% \\
    (20, 30] & 63.6\% & 63.6\% & \underline{56.8\%} & 63.6\% \\
    (30, 40] & 67.2\% & 67.3\% & \underline{58.2\%} & 67.3\% \\
    $(40, \infty)$ & 69.4\% & 68.8\% & \underline{56.8\%} & 68.8\%\\
    \hline
    \end{tabular}
    \caption{Fraction of divergence in the first half of the program, grouped by program length $L$. 
    In the column headers, H/R represents Hierarchical/Regular beam search under \textbf{Syntactic} constraint, and the number represents beam width $W$. 
    The column with the lowest fraction is underlined.
    }
    \label{tab:candidatevariation}
\end{table}

\subsection{Rejection by Constraints} \label{rejected}
In this section we give representative examples on what program candidates are rejected by our syntactic and symbol table constraints.

\paragraph{Syntactic Constraints} As mentioned in Section~\ref{sec:implementation}, about $26\%$ of the lines do not have pseudocode.
They may correspond to ``\texttt{\}}", ``\texttt{int main()\{}",  ``\texttt{\{}", "\texttt{return 0}", ``\texttt{\};}" or ``\texttt{;}".
These lines need contextual information to select valid code pieces and naïvely combining the top 1 candidate from each line independently will always produce grammatically invalid programs.
Syntactic constraints also rule out stylistic ambiguities. 
For example, when there is only one statement within an \texttt{if} statement, the programmer can optionally include a curly brace.
However, the pseudocode does not contain such detailed information about style.
Both ``\texttt{if(\dots)\{}" and ``\texttt{if(\dots)}" might be valid, but only one of them can be correct given the context of a program. 
Our syntactic constraints, which contain a curly brace constraint, can help us select the right code piece.

\paragraph{Symbol Table (SymTable) Constraints}
Pseudocode annotations are sometimes implicit about variable declarations. 
Given the instruction ``set N to 222222", both code pieces (1) ``\texttt{int N = 222222;}" and (2) ``\texttt{N = 222222;}" are potentially valid.
We might disambiguate this case with a SymTable constraint: if the variable is declared before in the same scope, then we know this code piece should not contain a repeated declaration and hence we should choose candidate (2); otherwise we should choose (1) to avoid using undeclared variables.
SymTable constraints are also helpful when the pseudocode does not put quotation marks around string/character literals.
Consider the instruction ``if lucky is A then do the following" with the ground truth code piece ``\texttt{if (lucky == 'A') \{}".
The model might misunderstand A as a variable name and generate ``\texttt{if (lucky == A) \{}".
This error can be ruled out by SymTable constraint if variable \texttt{A} is undeclared.

However, SymTable constraints do not preclude all errors related to declarations.
Consider the following generation where the last line is wrong:
% Consider using the `minted` package for nicer code snippets:
% https://www.overleaf.com/learn/latex/Code_Highlighting_with_minted
\begin{lstlisting}[language=C++, basicstyle=\small
]
int now = -1, cnt = 0;
for (int i = 0; i < n; ++i) {
    ... // some lines omitted
    // cnt = 1, now = v[i]; // gold
    int cnt = 1, now = v[i]; // pred
}
\end{lstlisting}
A programmer will usually not declare new variables in the last line of a variable scope.
However, technically this is not an invalid statement and the SymTable constraint fails to reject this wrong candidate.
Extra modelling is needed to take into account programming conventions and common sense.

\subsection{Code Piece Error Analysis} \label{beyondoracle}

So far we have focused on combining independent candidates from each line together to search for the target program. 
This heavily depends on the underlying model to generate potentially correct code pieces.
However, in 32\% of the programs at least one ``hard" line has no generated code piece that is functionally equivalent to the solution, thus indicating plenty of room for improvement.
To help the readers understand the bottleneck for code piece generation and point out important future directions, we randomly sampled 200 ``hard" lines and manually analyzed why the generation fails by looking at the top 1 candidate of the model.
The error analysis is available on our \href{https://github.com/ruiqi-zhong/SemanticScaffold/blob/master/hard_lines_category.csv}{GitHub}.
%\footnote{Our error analysis is available on \href{https://github.com/ruiqi-zhong/SemanticScaffold/blob/master/hard_lines_category.csv}{GitHub}.}

We group the failures into the following categories, giving a detailed breakdown and examples in Figure \ref{fig:categorized_error_analysis}.
(a) The model generation is wrong despite clear pseudocode; this typically happens when the gold code piece is long or highly compositional.
(b, c) The pseudocode contains ambiguity; the model generation is reasonable but either needs (b) variable type clarification or (c) syntactic context.
This requires incorporating contextual information of the program into the code piece generation process.
(d, e) The pseudocode either (d) consists of variable name typos or (e) is completely wrong.

\bibliography{acl2020}
\bibliographystyle{acl_natbib}
\appendix
\clearpage

\section{Appendices}
\label{sec:appendix}
\subsection{Primary Expressions}
Table \ref{tab:primary_expression_table} contains the grammar we use for the syntactic constraint and Table \ref{tab:terminals} defines the generation of terminal symbols.

\subsection{CFG Description Size of SymTable} \label{sec:CFGDescription}
We show that we cannot specify the SymTable constraint in a context free grammar without exponential description complexity w.r.t.\ the number of variables declared.
The intuition is that, since repeated declarations of a variable are not allowed, we need to keep track of all the variables that have been declared every time when verifying whether the next line is valid; however, a CFG, when transformed into a pushdown automata, is only allowed to peek at the top of the stack to decide the state transition. 
This means the symbol on the top of the stack, the state, or the transition rule need to have full information of about whether each variable has been declared, which contains exponentially many possibilities w.r.t.\ the number of variables.

Our proof is an adaptation of \citet{ellul2005regular}, which proves this property for the language that accepts all the permutations of a fixed number of variables.
We refer the readers to this paper if more details of the proof are needed.
To formalize, we consider a simple grammar of $K$ characters $\{ v_{1}, \dots, v_{K} \}$, where $v_{i}$ means, semantically, declaring the variable $v_{i}$, and the language $L$ consists of all the possible sequences of declarations that have no repetition.
\begin{equation}
    L = \{\mathrm{concat}_{j=1}^{k} v_{a_{j}} | a_{j_{1}} \neq a_{j_{2}} \text{ if } j_{1} \neq j_{2}, k \leq K\}
\end{equation}

We prove that 
\begin{theorem} \label{Description-complexity}
$L$ has at least $\Tilde{\Omega}(1.37^{K})$ description complexity\footnote{$\Tilde{\Omega}$ ignores all the poly($K$) multiplicative factors.} as a context free grammar.
\end{theorem}
Intuitively, it means if we want to use a CFG to specify $L$, we need the sum of total length of the production rules and number of symbols to be at least exponential.

\paragraph{Proof:} Since we can convert any CFG with size $B$ to Chomsky Normal Form (CNF) with size $O(B^{2})$, the above statement would be implied if we prove that $L$ needs $\Tilde{\Omega}(1.37^{2K}) = \Tilde{\Omega}(1.89^{K})$ description size in Chomsky Normal Form.

We use Lemma 31 from \citet{ellul2005regular}:
\begin{lemma}\label{symbolyield}
Let $S$ be the start symbol of the CFG. Then for all $w \in L $, there exists a symbol $A$ with %such that
\begin{equation}
    S \; \Longrightarrow^{*} \; \alpha A \beta \; \Longrightarrow^{*} \; w
\end{equation}
such that if $A$ yields $y$ in $w$ (i.e. $w = \alpha y \beta$), $\frac{1}{3}|w| \leq |y| \leq \frac{2}{3}|w|$.

In other words, for any member of the language, we can find a symbol in the derivation responsible for between 1/3 and 2/3 of the final yield.
\end{lemma}

Let $P_{K}$ be all sequences of permutations of the $K$ variables and thus $P_{K} \subset L$.
Then by Lemma \ref{symbolyield}, for every permutation $\pi \in P_{K}$ we can find yield $y_{\pi}$ that is yielded by a single symbol such that $\frac{1}{3}K \leq |y_{\pi}| \leq \frac{2}{3}K$.
Now we consider two permutations $\pi_{1}$ and $\pi_{2}$. If $y_{\pi_{1}}$ and $y_{\pi_{2}}$ are yielded by the same symbol, then they must have the same length (this is the part where the proof is slightly different from \citet{ellul2005regular}): suppose the contrary, w.l.o.g., let $|y_{\pi_{1}}| > |y_{\pi_{2}}|$.
By the definition of a context free grammar, we can replace the sub-string $y_{\pi_{2}}$ in $\pi_{2}$ by $y_{\pi_{1}}$ to create a new string $y'_{\pi_{2}}$ which is still a member of $L$. 
We have $|y'_{\pi_{2}}| = K - |y_{\pi_{2}}| + |y_{\pi_{1}}| > K$ by assumption.
However, there are in total $K$ variables; by the pigeonhole principle there must be a variable that is declared twice, and hence $y'_{\pi_{2}} \notin L$ and we obtain a contradiction.

Then all the assumption needed by Theorem 30 in \citet{ellul2005regular} hold and $L$ has description complexity $\Tilde{\Omega}(1.89^{K})$ in CNF and hence $L$ has description complexity $\Tilde{\Omega}(1.89^{K/2}) = \Tilde{\Omega}(1.37^{K})$. $\square$

\subsection{Hardness of Satisfying SymTable} \label{hardness}
We show that combining code pieces from each line under the SymTable constraint is NP-Hard in general.
We first remind the readers of the set packing problem:

\begin{definition}
Assume the universe to be $\mathcal{V}$, and suppose we are given a family of subsets $\mathcal{S}$ from the power set of $\mathcal{V}$, i.e.\ $P(\mathcal{V}) = \{S \mid S \subseteq \mathcal{V} \}$ and $\mathcal{S} \subseteq P(\mathcal{V})$.
We want to determine whether we can find a packing $\mathcal{K} \subseteq \mathcal{S}$ for which all sets in $\mathcal{K}$ are pairwise disjoint and with size $|\mathcal{K}| \ge L$ for some fixed $L > 0$.
This problem is called the set packing problem, and is known to be NP-complete.
\end{definition}

Following the notation in section \ref{sec:CFGDescription}, for each line $l \in [L]$, we construct the $C = |\mathcal{S}|$ code piece candidates $y_{lS}$ for $S \in \mathcal{S}$ as

\begin{equation}
    y_{lS} = \mathrm{concat}_{v\in S}v .
\end{equation}
We easily see that there is a set packing of size $L$ if and only if there is a valid code piece combination under SymTable constraint (declarations need to be disjoint for each line). 
Hence we finish our reduction proof. $\square$

\subsection{Beam Search on Unseen Problems}
Table \ref{tab:problem_beamsize} contains similar information as in Table \ref{tab:worker_beamsize}, except that the results are obtained on testing with unseen problems. 
The exact same conclusion holds: for regular beam search, small beam size hurts performance, but hierarchical beam search can solve this problem.

\begin{table}[]
    \centering
    \textbf{Test Against Unseen Problems, Syntactic}
    \begin{tabular}{cccc}
    \hline
    Method, Width & $B$=1 & $B$=10 & $B$=10$^{2}$  \\
        \hline
        H, $W$=10 & 27.4\% & 35.3\% & 42.0\% \\
        H, $W$=25 & 27.5\% &  35.4\% & 42.1\%\\
        H, $W$=50 & \textbf{27.5\%} & \textbf{35.4\%} & \textbf{42.1\%}\\
        R, $W$=200 & 27.1\% & 34.7\% & 41.0\%\\
        \hline
    \end{tabular}
    
    \textbf{Test Against Unseen Problems, SymTable}
    \begin{tabular}{cccc}
    \hline
        Method, Width & $B$=1 & $B$=10 & $B$=10$^{2}$  \\
        \hline
        H, $W$=10 & 30.3\% & 38.1\% & 43.1\% \\
        H, $W$=25 & 30.9\% &  39.2\% & 45.7\%\\
        H, $W$=50 & \textbf{31.0\%} & \textbf{39.2\%} & \textbf{45.9\%}\\
        R, $W$=200 & 30.7\% & 38.9\% & 45.4\%\\
        \hline
    \end{tabular}
    
    \caption{Comparison of different beam size with \textbf{Syntactic} and \textbf{SymTable} constraint when tested against unseen problems. H/R refers to hierarchical/regular beam search and $W$ is the beam width.
    This table is structured similarly as \ref{tab:worker_beamsize} .
    }
    \label{tab:problem_beamsize}
\end{table}

\subsection{Variation under SymTable Constraints}
Table \ref{tab:candidatevariation_SymTable} contains similar information as Table \ref{tab:candidatevariation}, but for SymTable constraints.
The same trend holds: regular beam search with small beam size have fewer variations in the first half of the program.

\begin{table}[]
    \centering
    \begin{tabular}{c|cccc}
    \hline
    Length $L$ & H 25 & H 50 & R 50 & R 200\\
      \hline
    (0, 10] & 40.7\% & 41.5\% & 39.4\% & 41.5\%  \\
    (10, 20] & 60.9\% & 59.8\% & 54.3\% & 61.3\% \\
    (20, 30] & 62.2\% & 61.3\% & 54.2\% & 61.3\% \\
    (30, 40] & 66.0\% & 66.1\% & 56.8\% & 66.1\% \\
    $(40, \infty)$ & 69.0\% & 68.7\% & 57.9\% & 68.7\%\\
    \hline
    \end{tabular}
    \caption{Fraction of divergence in the first half of the program, grouped by program length $L$. In the column headers, H/R represents Hierarchical/Regular beam search under \textbf{SymTable} constraint, and the number represents beam width $W$. 
    }
    \label{tab:candidatevariation_SymTable}
\end{table}

\begin{table*}[]
    \centering
    \begin{tabular}{|c|c|}
    \hline
    \textbf{Symbol} & \textbf{Production Rule} \\
    \hline
    program  & stmt program \\
      & function program \\
    \hline
    stmt & for\_stmt \\
      & if\_stmt \\
      & while\_stmt \\
      & dowhile\_stmt \\
      & terminal\_stmt ; \\
    \hline
    $X^{*}$ & $X^{*} X$  \\
     & $X$\\
     & $\langle$ EMPTY $\rangle$ \\
    \hline
    function & return\_type function\_name ( args) \{$_{\text{start}}$ stmt$^{*}$ \}$_{\text{end}}$ \\
     & return\_type function\_name ( type$^{*}$);\\
    \hline
    args & $\langle$ EMPTY $\rangle$ \\
    & arg , args\\
    \hline
    arg & type arg\_name \\
    \hline
    for\_stmt & for$_{\text{start}}$  terminal\_parentheses terminal\_stmt$_{\text{end}}$;\\
    & for$_{\text{start}}$ terminal\_parentheses \{stmt$^{*}$\}$_{\text{end}}$\\
    \hline
    while\_stmt & while$_{\text{start}}$  terminal\_parentheses terminal\_stmt$_{\text{end}}$;\\
    & while$_{\text{start}}$ terminal\_parentheses \{stmt$^{*}$\}$_{\text{end}}$\\
    \hline
    dowhile\_stmt & do$_{\text{start}}$ \{stmt$^{*}$\} while terminal\_parentheses$_{\text{end}}$;\\
    & do$_{\text{start}}$ terminal\_stmt while terminal\_parentheses$_{\text{end}}$;\\
    \hline
    if\_stmt & single\_if\_stmt elif\_stmt$^{*}$ else\_stmt \\
     & single\_if\_stmt elif\_stmt$^{*}$\\
     \hline
     single\_if\_stmt & if$_{\text{start}}$ terminal\_parentheses terminal\_stmt$_{\text{end}}$;\\
      & if$_{\text{start}}$ terminal\_parentheses \{stmt$^{*}$\}$_{\text{end}}$ \\
     \hline
    elif\_stmt & elif$_{\text{start}}$ terminal\_parentheses terminal\_stmt$_{\text{end}}$;\\
    & elif$_{\text{start}}$ terminal\_parentheses \{stmt$^{*}$\}$_{\text{end}}$\\
    \hline
    else\_stmt & else$_{\text{start}}$ terminal\_stmt$_{\text{end}}$;\\
    & else$_{\text{start}}$ \{stmt$^{*}$\}$_{\text{end}}$\\
    \hline
    \end{tabular}
    \caption{The full primary expression grammar we are using. 
    Each line is a production rule. $X$ is a generic symbol.}
    \label{tab:primary_expression_table}
\end{table*}

\begin{table*}[]
    \centering
    \begin{tabular}{|c|c|}
    \hline
    \textbf{Terminal} & \textbf{Implementation}  \\
    \hline
    terminal\_parentheses & a string that has matching parentheses and starts with parentheses \\
    \hline
    terminal\_stmt & a string that does not contain ``;", ``for",   ``if", ``else", ``while", ``do"\\
    \hline
    for, if, else, while, do & reserved key words \\
    \hline
    function\_name, arg\_name & function name and function argument name\\
    \hline
    return\_type, type & type in C++ \\
    \hline
    \end{tabular}
    \caption{The definition of the terminals appearing in Table 
    \ref{tab:primary_expression_table}}
    \label{tab:terminals}
\end{table*}

\end{document}